%% file: main.tex
\documentclass[runningheads]{llncs}

\usepackage{graphicx}
\usepackage{subcaption}
\usepackage{hyperref}       
\usepackage{url}            
\usepackage{booktabs}       
\usepackage{amsfonts}       
\usepackage{nicefrac}       
\usepackage{microtype}      
\usepackage{color}
\usepackage{float}
\usepackage{authblk}

\usepackage{enumitem}

\begin{document}
\title{Confidence intervals uncovered: \\ Are we ready for real-world medical imaging AI?}
\titlerunning{Confidence intervals uncovered}
%

\author{Evangelia Christodoulou\inst{1,2,3,\star} \and Annika Reinke\inst{1,4,\star} \and Rola Houhou\inst{1,3} \and Piotr Kalinowski\inst{1,3,5} \and Selen Erkan\inst{6} \and Carole H. Sudre\inst{7,8} \and Ninon Burgos\inst{9} \and Sofiène Boutaj\inst{9} \and Sophie Loizillon\inst{9} \and Maëlys Solal\inst{9} \and Nicola Rieke\inst{10} \and Veronika Cheplygina\inst{11} \and Michela Antonelli\inst{8,12} \and Leon D. Mayer\inst{1,3}  \and Minu D. Tizabi\inst{1,3}  \and M. Jorge Cardoso\inst{8} \and Amber Simpson\inst{13,14} \and Paul F. Jäger\inst{4,6} \and Annette Kopp-Schneider\inst{15}\and Gaël Varoquaux\inst{16,\star} \and Olivier Colliot\inst{9,\star} \and Lena Maier-Hein\inst{1,3,4,17,18,\star}}


 \renewcommand{\thefootnote}{\fnsymbol{footnote}}
 \footnotetext[1]{\tiny Shared first/last authors: E. Christodoulou and A. Reinke/L. Maier-Hein, O. Colliot, and G. Varoquaux}

  \authorrunning{E. Christodoulou et al.}


 \institute{German Cancer Research Center (DKFZ) Heidelberg, Div. Intelligent Medical Systems, Germany 
  \email{evangelia.christodoulou@dkfz-heidelberg.de} \and AI Health Innovation Cluster, Germany \and 
 National Center for Tumor Diseases (NCT), NCT Heidelberg, a partnership between DKFZ and Heidelberg University Hospital, Germany
 \and
   DKFZ Heidelberg, Helmholtz Imaging, Germany 
 \and
  HIDSS4Health - Helmholtz Information and Data Science School for Health, Germany 
  \and DKFZ Heidelberg, Interactive Machine Learning Group, Germany
  \and MRC Unit for Lifelong Health and Ageing at UCL and Centre for Medical Image Computing, Department of Computer Science, University College London, UK
   \and School of Biomedical Engineering and Imaging Science, King’s College London, UK
  \and Sorbonne Université, Institut du Cerveau - Paris Brain Institute - ICM, CNRS, Inria, Inserm, AP-HP, Hôpital de la Pitié-Salpêtrière, France 
   \and NVIDIA, Germany
   \and Department of Computer Science, IT University of Copenhagen, Denmark
 \and  Centre for Medical Image Computing, University College London, UK
 \and School of Computing, Queen’s University, Canada
 \and Department of Biomedical and Molecular Sciences, Queen’s University, Canada
  \and Division of Biostatistics, DKFZ,  Germany
 \and Parietal project team, INRIA Saclay-Île de France, France 
 \and Faculty of Mathematics and Computer Science, Heidelberg University, Germany
  \and Medical Faculty, Heidelberg University, Germany }



 
\maketitle              
\begin{abstract}
Medical imaging is spearheading the AI transformation of healthcare. Performance reporting is key to determine which methods should be translated into clinical practice. Frequently, broad conclusions are simply derived from mean performance values. In this paper, we argue that this common practice is often a misleading simplification as it ignores performance variability. Our contribution is threefold. (1) Analyzing all MICCAI segmentation papers (n = 221) published in 2023, we first observe that more than 50\% of papers do not assess performance variability at all. Moreover, only one (0.5\%) paper reported confidence intervals (CIs) for model performance. (2) To address the reporting bottleneck, we show that the unreported standard deviation (SD) in segmentation papers can be approximated by a second-order polynomial function of the mean Dice similarity coefficient (DSC). Based on external validation data from 56 previous MICCAI challenges, we demonstrate that this approximation can accurately reconstruct the CI of a method using information provided in publications. (3) Finally, we reconstructed 95\% CIs around the mean DSC of MICCAI 2023 segmentation papers. The median CI width was 0.03 which is three times larger than the median performance gap between the first and second ranked method. For more than 60\% of papers, the mean performance of the second-ranked method was within the CI of the first-ranked method. We conclude that current publications typically do not provide sufficient evidence to support which models could potentially be translated into clinical practice.

\end{abstract}

\section{Introduction}\label{sec:intro}
As demonstrated by the fact that more than 530 of the first 692 AI in healthcare products approved by the U.S. Food and Drug Administration (FDA) fall within the application domain of medical imaging \cite{us2022artificial}, medical imaging is spearheading the AI-powered transformation of healthcare. Performance reporting and comparisons of medical imaging models are key to determining their potential for clinical translation. Clinical translation requires approval by regulatory agencies such as the U.S. FDA, whose  recommendations insist on the importance of characterizing variability and reporting confidence intervals (CIs); for instance in \cite{food2019recommended,21cfr8922070,us2007statistical,21cfr8922060}). A recent paper \cite{chen2023regulatory}  written by FDA staff describes regulatory science principles on performance assessment of AI algorithms in imaging and emphasizes that \textit{"The statistical analysis plays a critical role in the assessment of machine learning (ML) performance but may be under-appreciated by many ML developers"}. Current practice in reporting results (including that of the authors!) often does not fulfill these requirements and thus far from lends itself to determining whether a medical imaging model is suited for clinical translation. The underlying question is: Can we really trust performance claims made in publications? The purpose of this paper was to address this important question:
\begin{enumerate}[itemsep=0pt, parsep=0pt, topsep=0pt, partopsep=0pt]
    \item Based on a comprehensive analysis of all MICCAI 2023 segmentation papers, we show that performance variability is rarely accounted for in the medical image analysis community.
    \item To demonstrate the implications of the reporting bottleneck, we propose a work-around to approximate variability parameters from the information provided in publications. Specifically, we show that the unreported standard deviation (SD) in segmentation papers can be approximated using a second-order polynomial function of the mean Dice similarity coefficient (DSC).  
    \item Using our proposed approximation method, we reconstruct CIs for the published MICCAI papers and provide evidence that the praise proposed methods receive is often not supported by sufficient evidence.
\end{enumerate}
\begin{figure}[H]
    \centering
    \includegraphics[width=0.95\textwidth]{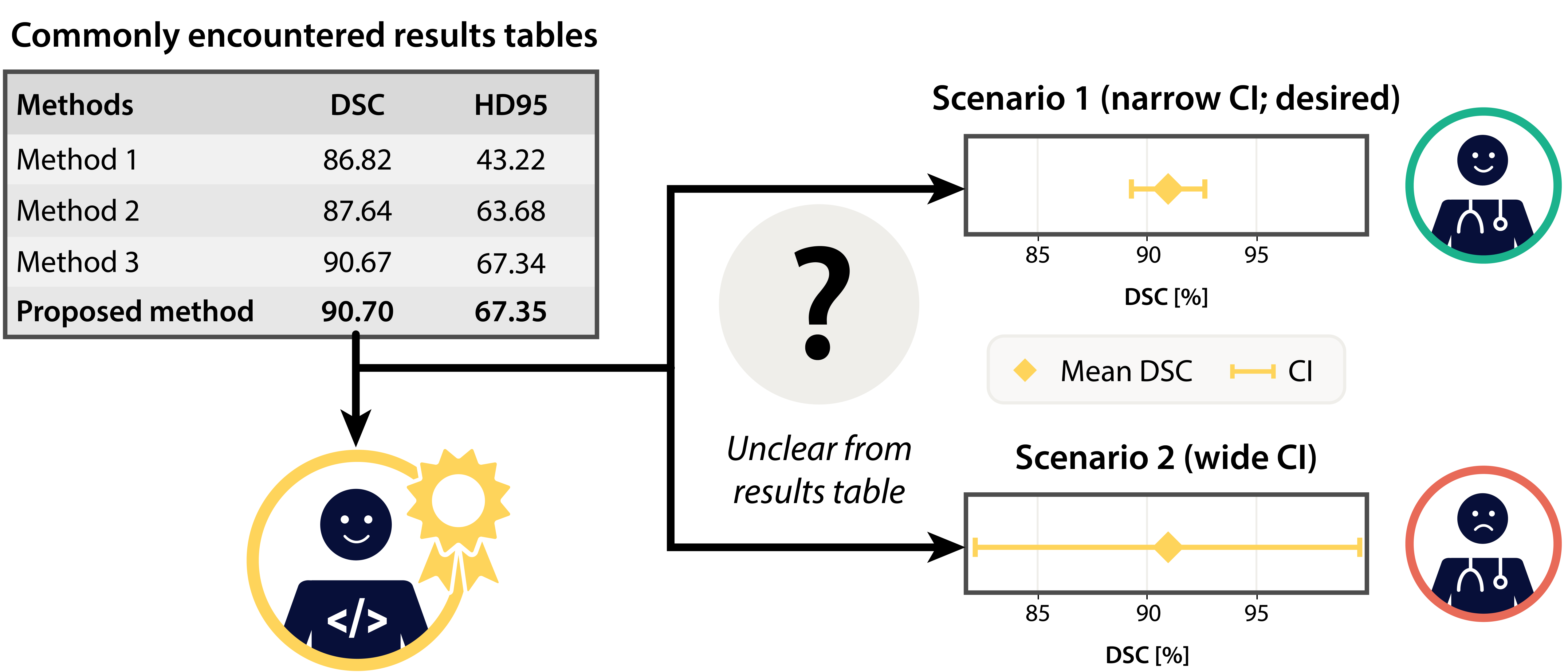}
    \caption{Common practice in medical imaging algorithm performance reporting leaves many open questions.}
    \label{fig:fig1}
\end{figure}

\begin{figure}[h]
    \centering
    \includegraphics[width=1\textwidth]{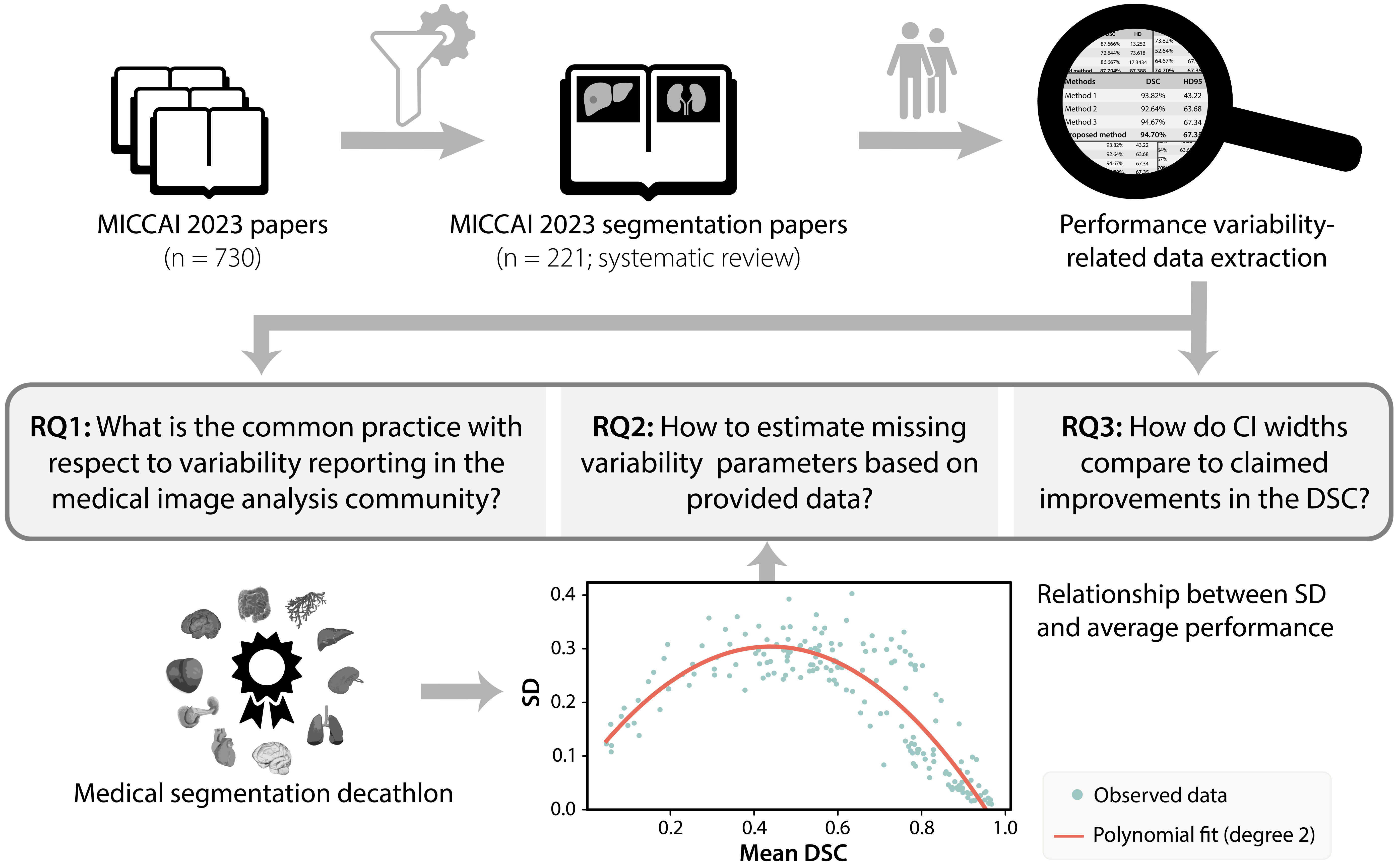}
    \caption{\textbf{Top: Research Questions (RQs) investigated.} Bottom: Model-based approximation of the SD. The observed (light green) and approximated (red line) SD is shown as a function of the mean DSC using a polynomial fit. Each point corresponds to the values of mean DSC and observed SD in the Medical Segmentation Decathlon data.}
    \label{fig:rqs}
\end{figure}
\section{Methods}\label{sec:methods}
Assessment of AI model performance variability is crucial as it directly impacts the model’s reliability in clinical practice. While variability reporting guidelines---in particular regarding the inclusion of CIs---are available in the clinical prediction modeling domain \cite{collins2024tripod+}, such practices are still unfamiliar in the medical imaging domain. 

In this paper, we focus on two statistical concepts capturing performance variability:
SD is a measure of the dispersion or spread of data points from the mean value. For example, given a set of performance metric values (e.g., DSC values of a model on multiple images) the SD states how much these values vary from the average performance. A small SD indicates that the values are close to the mean, while a large one suggests that the values are more dispersed. A CI can be used to estimate the range within which a population parameter (such as the mean) is expected to lie with a certain level of confidence. For example, a 95\% CI for the mean suggests that if we were to take many samples and calculate the CI for each, about 95\% of these intervals would contain the true population mean. CIs provide a measure of the precision of an estimate. Their widths approach 0 for infinite sample sizes.
	
To identify current practices in performance variability reporting and further raise awareness on this matter in the medical imaging community, our work addresses research questions (RQs) depicted in Figure~\ref{fig:rqs}. 

\subsection{Systematic review of MICCAI 2023 segmentation papers}\label{ssec:methods-review}
Given that segmentation is a key focus of MICCAI and DSC is the community's primary metric \cite{maier2018rankings}, our study concentrates on segmentation papers. From each of the identified segmentation papers we extracted information on the claims of the paper, method performance, its variability, and validation practices. To reduce the bias in extracting information from the papers, each paper was screened independently by two researchers. Subsequently, three additional researchers, distinct from those involved in data extraction, addressed data extraction conflicts. With the inclusion criteria being use of a test set for validation and mention of the exact test set size and mean DSC values, we identified all segmentation papers for which we could approximate the SD and CI. We excluded papers that solely used a random train/test split with no validation set (because there is a risk that the test set was used for validation, e.g., for model selection, leading to overoptimistic performance estimates) or only provided performance information graphically.

\begin{figure}[h]
    \centering
    \includegraphics[width=0.9\textwidth]{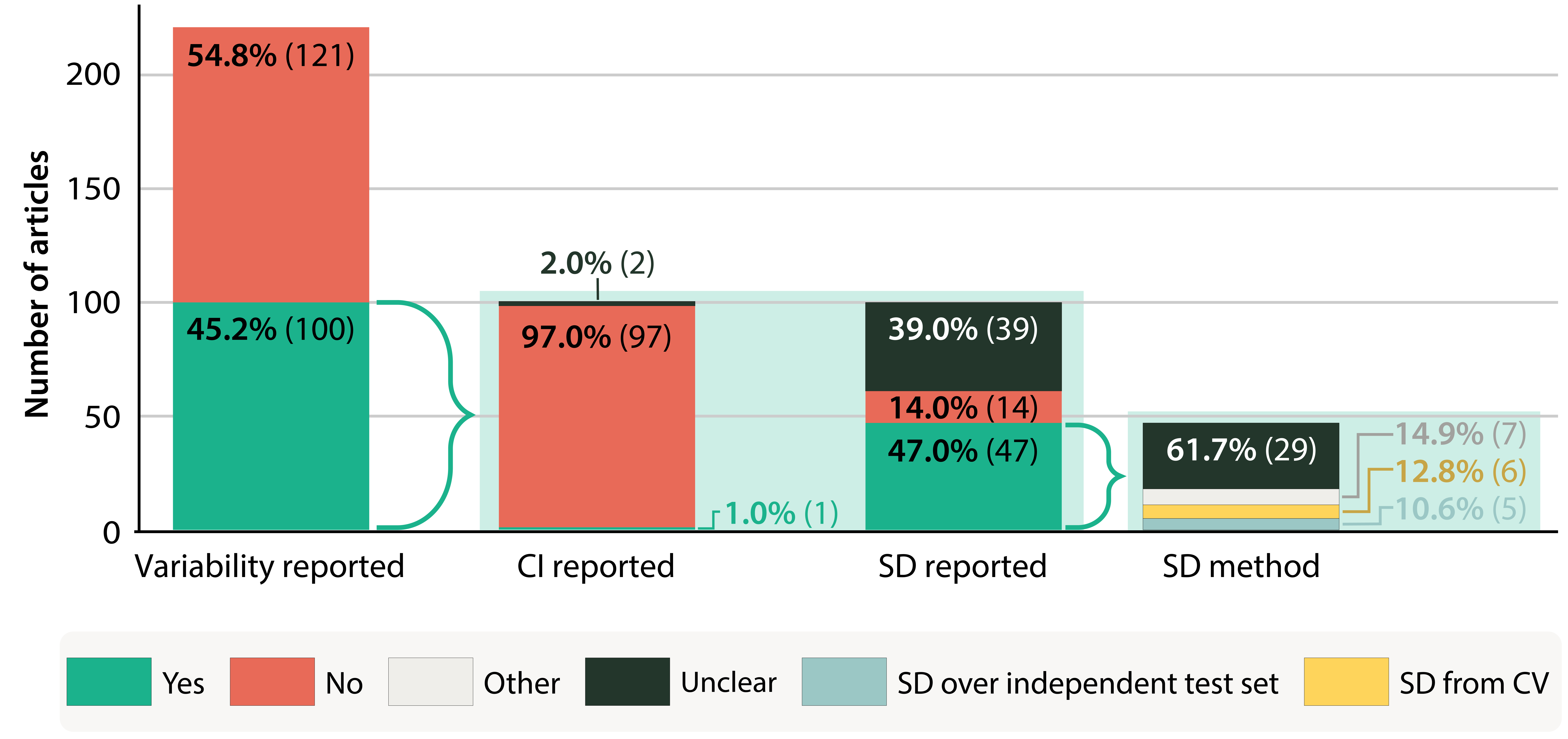}
    \caption{\textbf{Manner of reporting performance variability in all n = 221 MICCAI 2023 segmentation papers.} CI: confidence Interval; SD: standard deviation; CV: cross-validation.}
    \label{fig:descriptive-variability}
\end{figure}

\subsection{Approximation of missing variability parameters }\label{ssec:methods-approx}
To develop a method for approximating missing SD and CI from data present in publications, we used data from the Medical Segmentation Decathlon challenge \cite{antonelli2022medical,simpson2019large}, which saw 19 models competing on 10 different segmentation tasks in different anatomical regions. Stratifying by task, we calculated both the mean and SD of the DSC for each model of the challenge’s test set resulting in 189 SD values. When plotting the values of mean DSC against the SD, it seemed that a 
\noindent second-order polynomial curve would fit the data points reasonably well (see Figure~\ref{fig:rqs} bottom). To formally address this functional relationship, using the \texttt{statsmodel} library in Python, we fitted a generalized linear model (GLM) with a log link function, assuming a Gamma distribution for SD, which was expressed as a function of a second-order polynomial of mean DSC. The final model was $Log(SD)=2.0310 + 0.0726\cdot DSC_\mu -0.0008\cdot DSC_\mu^2$. Following compensation for missing SD values, we computed CIs around their respective mean DSC using a parametric approach, as we had no access to the test data and models used in the papers, and could thus not use bootstrapping methods. The results from this parametric approach have been shown to closely approximate those from a non-parametric approach that uses bootstrapping \cite{jurdi2023confidence}, justifying our choice of method. For calculation of CI, we used $\left[ DSC_\mu - t_{n-1, 1-\alpha/2} \cdot \frac{SD}{\sqrt{n}}, DSC_\mu + t_{n-1, 1-\alpha/2} \cdot \frac{SD}{\sqrt{n}} \right]$, with $DSC_\mu$ being the mean DSC, $n$ the test size, $t_{n-1, 1-\alpha/2}$ the quantile of the t distribution, $n-1$ the degrees of freedom and $\alpha$ the level of significance. We set $\alpha$ to 0.05, corresponding to 95\% CIs.

\section{Experiments and Results}\label{sec:results}
\paragraph{\textbf{RQ1: Common practice with respect to variability reporting}} \hfill \\
From all 730 papers published in the scope of MICCAI 2023, we identified 221 (30.3\%) segmentation papers. As shown in Figure~\ref{fig:descriptive-variability}, more than half of the papers (54.8\%) did not report any kind of variability. CIs were reported in only one paper (0.5\%). Of those that did report variability, only 47\% reported SD (21\% of all segmentation papers), and only 5\% combined SD with a graphical representation of variability. However, for 61.7\% of the papers that reported SD, its method of computing was not specified. 83.3\% of papers claimed that they outperformed the state of the art.

\begin{figure}[t]
    \centering
    \includegraphics[width=\textwidth]{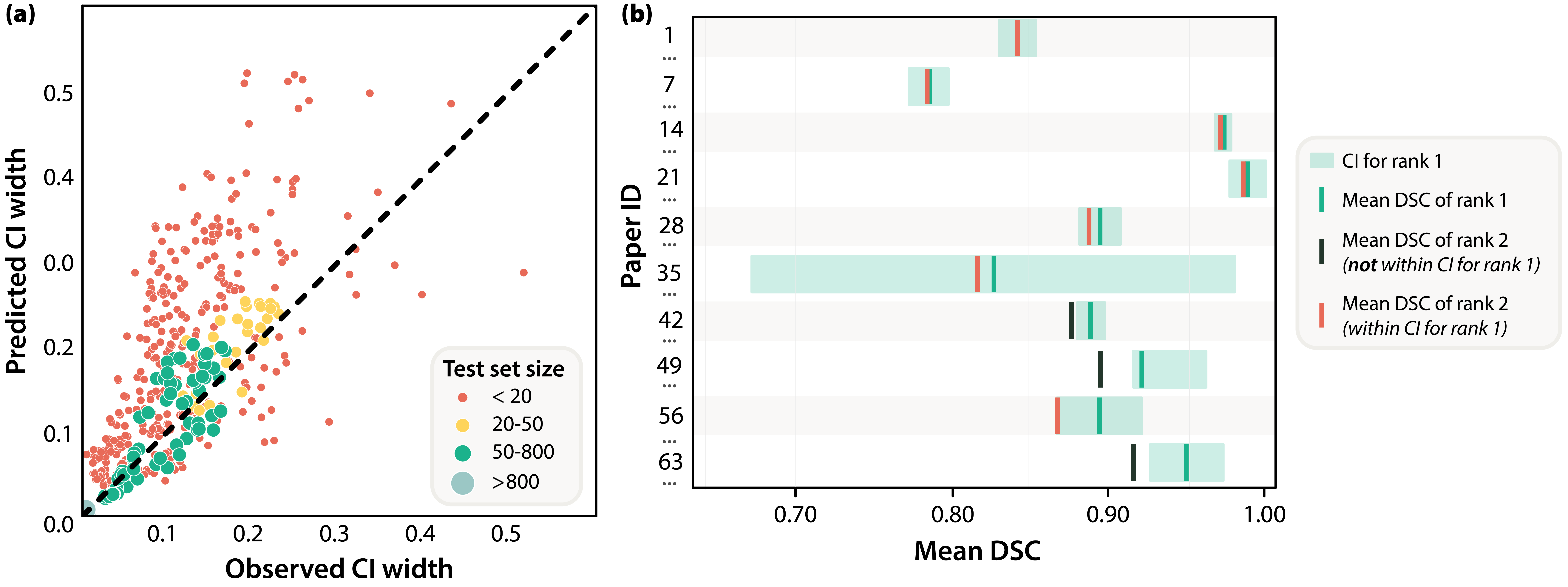}
    \caption{(a) Calibration plot of predicted CI widths (using approximated SD values from our model) versus observed CI widths (using the actual values of SD in the data) using data from 56 past MICCAI segmentation challenges. (b) Graph demonstrating if the mean DSC of the second-ranked method was within the DSC CI of the first-ranked method, including every seventh of the papers meeting our inclusion criteria.}
    \label{fig:calibration-ci-mean}
\end{figure}

\paragraph{\textbf{RQ2: Quality of SD approximation}} \hfill \\
The quality of our polynomial fit on the development data is illustrated in Figure~\ref{fig:rqs} (bottom).  
For external validation of our data imputation method, we obtained access to the performance data from 56 past MICCAI segmentation challenges \cite{maier2018rankings}, comprising results for 213 different methods applied to 124 different tasks. For these, we computed the SD and CI both from the observed data and with our approach. According to our results, the proposed model generalizes well to unseen data with a median (interquartile range (IQR)) difference between the observed and predicted CI width of 0.0024 (0.0097, 0.0422) for the dataset sizes $>$ 20 (better for increasing test set size as shown in Figure~\ref{fig:calibration-ci-mean}(a)).

\paragraph{\textbf{RQ3 Performance differences versus widths of CI}} \hfill \\
A total of 77 papers met our inclusion criteria for imputing the CI. The median/max (IQR) CI width for the first-ranked method was 0.03/0.31 (0.02, 0.06). The median (IQR) difference in mean DSC between the first- and second-ranked method (from now on referred to as \textit{delta DSC}) was 0.01 (0.00, 0.03). Thus, the median width of the CIs of the first-ranked method was about three times larger 
than the median delta DSC (see Figure~\ref{fig:boxplots} in the Supplement). For 64.9\% of papers, the mean performance of the second-ranked method was within the CI range of the first-ranked method (Figure~\ref{fig:calibration-ci-mean}(b)). The code for our experiments is available at: \url{https://github.com/IMSY-DKFZ/CI_uncovered} 

\section{Discussion}\label{sec:discussion}
Our work is the first to systematically analyze common practice with respect to model performance variability reporting in the field of medical image analysis. Our study clearly shows that reporting of performance variability, in particular reporting of CIs, is a rare exception. These reporting practices are at odds with the MICCAI reproducibility checklist guidelines \cite{miccaiguidelines} which include the following item: \textit{"A description of results with central tendency (e.g., mean) \& variation (e.g., error bars)"}. Even when variability is reported, for instance in the form of the SD, conclusions are commonly drawn based on mean performance values without taking variability into account. This reporting practice can be very misleading and is highly unhelpful in reaching the ultimate goal of clinical translation of imaging models. A model exhibiting a high mean metric score but large variability in performance may not be suitable for safety-critical real-world applications, in which low performance on even some images may have dramatic consequences for patients. 

A limitation of our study could be seen in the fact that we had to approximate the SDs and subsequently CIs based on the data available. However,  our external validation of our SD approximation  indicates high reliability (Figure~\ref{fig:calibration-ci-mean}(a)).

Related work on the topic of variability analysis is sparse. In an analysis of biomedical image analysis challenge reporting, \cite{wiesenfarth2021methods} showed that claims are often solely drawn from aggregated results in tables, which supports our hypothesis. \cite{jurdi2023confidence} likewise emphasize the lack of reporting CIs in medical image segmentation. Reporting on the Conference on Neural Information Processing Systems (NeurIPS) reproducibility program, \cite{pineau2021improving} state that \textit{"it seems surprising to have 87\% of papers that see value in clearly defining the metrics and statistics used, yet 36\% of papers judge that error bars are not applicable to their results"}. While our current analysis focuses on the variability of the \textit{trained model} (i.e., accounting from variance coming from the test set), \cite{bouthillier2021accounting} analyzed the variability of the \textit{learning procedure} (i.e., accounting for other sources of variances such as random seeds or hyperparameters) during initial method development. Additionally, \cite{maier2018rankings} investigated rankings in biomedical challenges and found that these are often not stable. Similarly, \cite{varoquaux2022machine} found that rankings between private and public leaderboards in Kaggle competitions were not stable.

In this paper, we focused on the question: Can we trust the reported mean performance results? Note that this is conceptually different from asking whether one method truly  outperforms another, as investigated in \cite{maier2018rankings}. In fact, the confidence in reported mean performance values, as measured by CIs, is necessary (but, of course, not sufficient) for deciding on whether a proposed algorithm is ready for clinical translation. Our study revealed that claims of scientific progress are typically based on small differences (around 0.01) in the mean DSC, suggesting that these were considered clinically relevant by the authors. In contrast, CIs are---on average---much wider. We consider this contradictory because if a difference of 0.01 matters, then, shouldn’t a CI with a much larger width be concerning (and at least be reported), as it means that the true mean may be substantially smaller than the reported one? 

Future work should not only be directed to fostering better reporting practices, but also address a complementary question: Does a published method really make an improvement over the state of the art? P-values are probably the most visible statistical tool in this context, yet, the standard view of statistical testing (null-hypothesis significance testing) is often considered as insufficient evidence both in machine learning \cite{benavoli2017time,bouthillier2021accounting} and in medical evaluation \cite{cleophas2008clinical}. One of their drawbacks is that a sufficiently large sample size can make any two models significantly different. However, a difference can be statistically significant but so small that it is clinically meaningless. To assess clinically relevant benefits,  "superiority margins"---as used in superiority testing for clinical trials---are an interesting concept that could  easily be adapted to CIs by adding a boundary \cite{cleophas2008clinical}.

In conclusion, we showed that current publications in the medical image analysis community typically do not provide sufficient evidence to support which models could potentially be translated into clinical practice. We hope that our results will trigger a major community shift towards uncertainty-aware performance assessment of medical image analysis models.

\begin{credits}
\subsubsection{\ackname} We acknowledge funding from the French government under management of Agence Nationale de la Recherche as part of the “Investissements d’avenir” program, reference ANR-19-P3IA-0001 (PRAIRIE 3IA Institute), reference ANR-10-IAIHU-06 (Agence Nationale de la Recherche-10-IA Institut Hospitalo-Universitaire-6) and reference ANR-23-CE17-0054.
This publication was further supported through state funds approved by the State Parliament of Baden-Württemberg for the Innovation Campus Health + Life Science Alliance Heidelberg Mannheim. Moreover, it received funding from the European Research Council (ERC) under the European Union’s Horizon 2020 research and innovation program (grant agreement no. 101002198, NEURAL SPICING). Part of this work was also funded by Helmholtz Imaging (HI), a platform of the Helmholtz Incubator on Information and Data Science.

\subsubsection{\discintname}
We declare the following competing interests: N. Rieke is an employee of NVIDIA. The remaining authors have no competing interests.
\end{credits}

%
%
\bibliographystyle{splncs04}
\bibliography{references}

 \include{supplementary}

\end{document}

%% file: supplementary.tex
\section*{Supplementary Material}\label{sec:suppl}


\begin{figure}
    \centering
    \includegraphics[width=\textwidth]{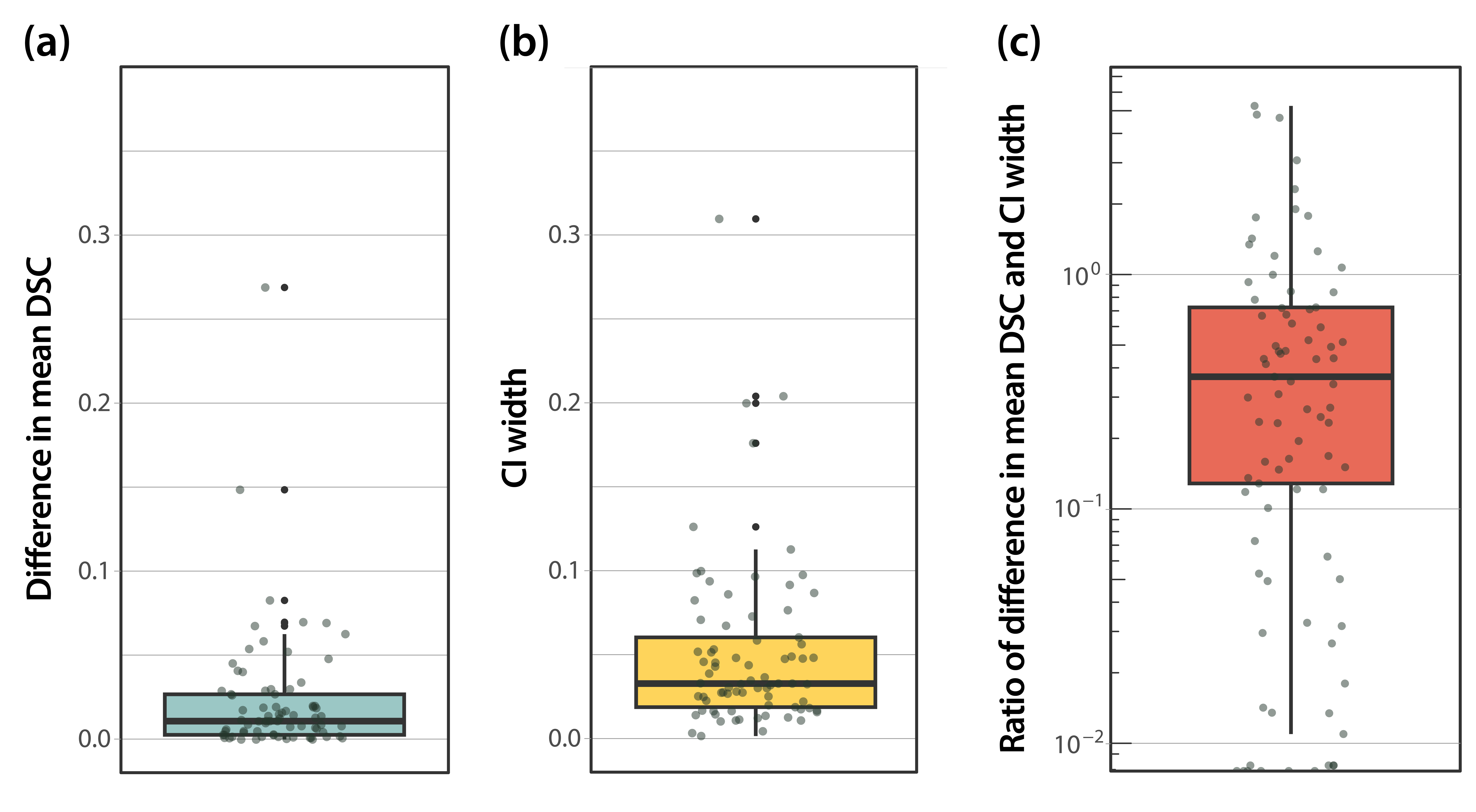}
    \caption{\textbf{The width of confidence intervals (CIs) is mostly larger than the performance gain.} Boxplots describing: (a) the difference in mean Dice Similarity Coefficient (DSC) between the two top-ranked methods within a paper, (b) the CI width of the first-ranked method of each paper, and (c) the ratio of difference in mean DSC for the two top-ranked methods methods within a paper and CI width of the first-ranked method of each paper.}
    \label{fig:boxplots}
\end{figure}

%% file: main.bbl
\begin{thebibliography}{10}
\providecommand{\url}[1]{\texttt{#1}}
\providecommand{\urlprefix}{URL }
\providecommand{\doi}[1]{https://doi.org/#1}

\bibitem{antonelli2022medical}
Antonelli, M., Reinke, A., Bakas, S., Farahani, K., Kopp-Schneider, A., Landman, B.A., Litjens, G., Menze, B., Ronneberger, O., Summers, R.M., et~al.: The medical segmentation decathlon. Nature communications  \textbf{13}(1), ~4128 (2022)

\bibitem{benavoli2017time}
Benavoli, A., Corani, G., Dem{\v{s}}ar, J., Zaffalon, M.: Time for a change: a tutorial for comparing multiple classifiers through bayesian analysis. The Journal of Machine Learning Research  \textbf{18}(1),  2653--2688 (2017)

\bibitem{bouthillier2021accounting}
Bouthillier, X., Delaunay, P., Bronzi, M., Trofimov, A., Nichyporuk, B., Szeto, J., Mohammadi~Sepahvand, N., Raff, E., Madan, K., Voleti, V., et~al.: Accounting for variance in machine learning benchmarks. Proceedings of Machine Learning and Systems  \textbf{3},  747--769 (2021)

\bibitem{21cfr8922060}
{CFR}: 21 {CFR} 892.2060 radiological computer-assisted diagnostic software for lesions suspicious of cancer (2020), \url{https://www.ecfr.gov/current/title-21/section-892.2060} [Accessed: 4 March 2024]

\bibitem{21cfr8922070}
{CFR}: 21 {CFR} 892.2070 medical image analyzer (2020), \url{https://www.ecfr.gov/current/title-21/section-892.2070} [Accessed: 4 March 2024]

\bibitem{chen2023regulatory}
Chen, W., Krainak, D., Sahiner, B., Petrick, N.: A regulatory science perspective on performance assessment of machine learning algorithms in imaging. Machine Learning for Brain Disorders pp. 705--752 (2023)

\bibitem{cleophas2008clinical}
Cleophas, G.C., Cleophas, M.T., Cleophas, T.J.: Clinical trials: Superiority-testing. Clinical Research and Regulatory Affairs  \textbf{25}(1),  31--39 (2008)

\bibitem{collins2024tripod+}
Collins, G.S., Moons, K.G., Dhiman, P., Riley, R.D., Beam, A.L., Van~Calster, B., Ghassemi, M., Liu, X., Reitsma, J.B., Van~Smeden, M., et~al.: Tripod+ ai statement: updated guidance for reporting clinical prediction models that use regression or machine learning methods. bmj  \textbf{385} (2024)

\bibitem{jurdi2023confidence}
Jurdi, R.E., Varoquax, G., Colliot, O.: Confidence intervals for performance estimates in 3d medical image segmentation. arXiv preprint arXiv:2307.10926  (2023)

\bibitem{maier2018rankings}
Maier-Hein, L., Eisenmann, M., Reinke, A., Onogur, S., Stankovic, M., Scholz, P., Arbel, T., Bogunovic, H., Bradley, A.P., Carass, A., et~al.: Why rankings of biomedical image analysis competitions should be interpreted with care. Nature communications  \textbf{9}(1), ~5217 (2018)

\bibitem{miccaiguidelines}
{MICCAI}: {MICCAI} reproducibility checklist (2021), \url{https://miccai2021.org/files/downloads/MICCAI2021-Reproducibility-Checklist.pdf} [Accessed: 7 March 2024]

\bibitem{pineau2021improving}
Pineau, J., Vincent-Lamarre, P., Sinha, K., Larivi{\`e}re, V., Beygelzimer, A., d'Alch{\'e} Buc, F., Fox, E., Larochelle, H.: Improving reproducibility in machine learning research (a report from the neurips 2019 reproducibility program). The Journal of Machine Learning Research  \textbf{22}(1),  7459--7478 (2021)

\bibitem{simpson2019large}
Simpson, A.L., Antonelli, M., Bakas, S., Bilello, M., Farahani, K., Van~Ginneken, B., Kopp-Schneider, A., Landman, B.A., Litjens, G., Menze, B., et~al.: A large annotated medical image dataset for the development and evaluation of segmentation algorithms. arXiv preprint arXiv:1902.09063  (2019)

\bibitem{us2007statistical}
{US Food and Drug Administration}, et~al.: Statistical guidance on reporting results from studies evaluating diagnostic tests. Rockville, MD: US FDA  (2007)

\bibitem{food2019recommended}
{US Food and Drug Administration}, et~al.: Recommended content and format of non-clinical bench performance testing information in premarket submissions: guidance for industry and food and drug administration staff (2019)

\bibitem{us2022artificial}
{US Food and Drug Administration}, et~al.: Artificial intelligence and machine learning ({AI}/{ML})-enabled medical devices. AI/ML-Enabled Medical Devices  (2022), \url{https://www.fda.gov/medical-devices/software-medical-device-samd/artificial-intelligence-and-machine-learning-aiml-enabled-medical-devices?trk=article-ssr-frontend-pulse_little-text-block} [Accessed: 4 March 2024]

\bibitem{varoquaux2022machine}
Varoquaux, G., Cheplygina, V.: Machine learning for medical imaging: methodological failures and recommendations for the future. NPJ digital medicine  \textbf{5}(1), ~48 (2022)

\bibitem{wiesenfarth2021methods}
Wiesenfarth, M., Reinke, A., Landman, B.A., Eisenmann, M., Saiz, L.A., Cardoso, M.J., Maier-Hein, L., Kopp-Schneider, A.: Methods and open-source toolkit for analyzing and visualizing challenge results. Scientific reports  \textbf{11}(1), ~2369 (2021)

\end{thebibliography}
